\documentclass[letterpaper, 10 pt, conference]{ieeeconf} 
\usepackage{graphicx}
\usepackage{pifont}
\usepackage{hyperref}
\usepackage[ruled]{algorithm2e}
\usepackage{multirow} 
\usepackage{booktabs}
\usepackage[caption=false]{subfig}
\usepackage{color} 
\usepackage{amsmath,amssymb,amsfonts} 
\usepackage{ulem}
\usepackage[short]{optidef}
\usepackage{cite}
\hypersetup{
    colorlinks=true,
    linkcolor=blue,
    citecolor=blue,
    urlcolor=blue,
}

\IEEEoverridecommandlockouts

\title{
MoMa-MO: LLM-Guided Mobile Manipulation Among Movable Obstacles with Attribute-Grounded Obstacle Memory}

\author{Qijin Song, and Weibang Bai$^{\ast}$
\thanks{This work was supported by the Key Laboratory of Intelligent Perception and Human-Machine Collaboration (ShanghaiTech University), Ministry of Education. Corresponding author: Weibang Bai {\itshape (E-mail: wbbai@shanghaitech.edu.cn).}}
\thanks{Q. Song and W. Bai are with the School of Information Science and Technology, ShanghaiTech University,  Shanghai 201210, China. }
}
\begin{document}

\maketitle
\thispagestyle{empty}
\pagestyle{empty}

\begin{abstract}
Mobile manipulation in cluttered environments often encounters blocked access to target objects or placement locations, leading to long detours or task failure. Pushing these obstacles offers an alternative, but choosing between pushing and taking a detour requires assessing whether the robot can move them. Semantic and geometric attributes provide initial cues, yet unobserved properties such as loading and friction can cause an apparently movable obstacle to resist pushing. To address this challenge, we introduce a large language model (LLM)-guided framework for Mobile Manipulation among Movable Obstacles (MoMa-MO). The framework incorporates Attribute-Grounded Obstacle Memory (AGOM) to associate semantic and geometric attributes with movability estimates and pushing outcomes in persistent obstacle records. By combining these records with current task and route context, the LLM plans navigation, obstacle pushing, and target manipulation, and revises its plan when execution feedback indicates an ineffective push. Experiments on a quadruped manipulator demonstrate correction of initial movability judgments during repeated observations, adaptation after ineffective pushes, and an overall task success rate of 88\% across five obstacle instances. In a matched comparison with an available detour, the framework maintains the same observed task success rate as navigation-only execution while reducing mean completion time by 31.8\% and mean base travel distance by 48.9\% over successful trials.
\end{abstract}

\section{Introduction}
\label{sec:introduction}

Mobile manipulators combine navigation and manipulation to perform tasks in complex environments~\cite{jiao2025int,burgess2024react}. During task execution, surrounding obstacles may block access to a target object or placement location. Approaches that treat these obstacles as fixed constraints rely on collision-free navigation to avoid them~\cite{yan2025dyn,nav2025placenet}, potentially requiring long detours or leaving the task inaccessible. When a blocking obstacle can be physically displaced, interaction offers an alternative way to obtain access or shorten the route~\cite{bi2025interactive,stilman2005navigation}. This motivates Mobile Manipulation among Movable Obstacles (MoMa-MO), in which obstacle interaction is incorporated into the execution of a manipulation task.

Quadruped manipulators combine legged mobility with arm-based interaction~\cite{lopes2023review, liu2026gaze}, providing the capabilities to navigate around obstacles or push them aside before continuing target manipulation. Coordinating these actions requires deciding when pushing offers a useful alternative to a detour. This decision depends on the available routes, the feasibility of approaching and contacting the obstacle, and whether the obstacle is expected to move under the robot's pushing action. Semantic and geometric attributes provide useful cues for this assessment~\cite{wang2020affordance}, but these cues must be interpreted together with the access requirements of the target task.

\begin{figure}[t]
    \centering
    \includegraphics[width=\linewidth]{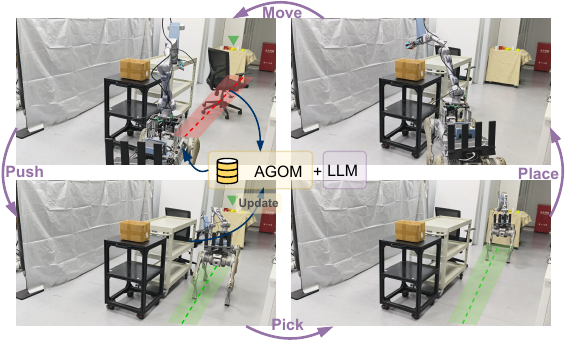}
    \caption{Mobile manipulation among movable obstacles with a quadruped manipulator. AGOM stores and updates obstacle information to support LLM-based task planning. The illustrated task combines move, push, pick, and place actions, with obstacle pushing opening access for subsequent target manipulation.}
    \label{fig:cover}
\end{figure}

Large language models (LLMs) offer semantic knowledge and task-level reasoning capabilities for interpreting scene information and coordinating robot actions~\cite{liu2025dynamem,qiu2025wildlma}. In MoMa-MO, these capabilities support initial movability assessment from obstacle attributes and comparison of pushing and navigation alternatives in the current task context. However, visual observations do not fully reveal physical properties such as loading and friction. A box initially judged movable may therefore resist pushing because of unobserved contents. The resulting interaction provides evidence for revising the initial estimate, but this evidence can inform later decisions only if it remains associated with the corresponding obstacle. Without such a persistent association, planning from current observations may repeat an ineffective push despite the previous outcome. Effective obstacle handling therefore requires retaining physical interaction evidence and reusing it as task execution progresses.

To address this challenge, this work presents an LLM-guided MoMa-MO framework centered on Attribute-Grounded Obstacle Memory (AGOM). AGOM maintains persistent obstacle records that associate semantic and geometric attributes with movability estimates and interaction histories. Relevant records are combined with current task and route context to support symbolic instruction planning, while dedicated modules validate and execute the resulting commands. Post-push observations update the corresponding obstacle record and inform plan revision when necessary. Obstacle displacement and route clearance are assessed separately to determine the interaction outcome and whether execution can proceed. Fig.~\ref{fig:cover} illustrates a task in which this framework coordinates obstacle pushing and subsequent target manipulation.

The main contributions of this work are:
\begin{itemize}

    \item We propose an LLM-guided MoMa-MO framework that coordinates navigation, obstacle pushing, and target manipulation through symbolic instruction planning, feasibility validation, and execution feedback.

    \item We introduce Attribute-Grounded Obstacle Memory (AGOM), a persistent obstacle-centered representation that supports movability initialization from semantic and geometric attributes and subsequent updates using physical interaction feedback.

    \item We evaluate the framework on a real quadruped manipulator, assessing movability classification, task execution across obstacle instances, and adaptation after ineffective pushes. In a matched comparison with an available detour, the framework reduces mean completion time and base travel distance over successful trials while maintaining the same observed task success rate as navigation-only execution.

\end{itemize}

\section{Related Work}
\label{sec:related_work}

\subsection{Quadruped Mobile Manipulation}

Quadruped manipulators combine legged locomotion~\cite{guo2025learning,li2026s} with arm-based interaction, extending mobile manipulation to environments that challenge wheeled platforms~\cite{lopes2023review}. 
Existing studies investigate whole-body control, coordinated locomotion and manipulation, and perception-driven interaction for tasks such as reaching, grasping, and object relocation~\cite{sombolestan2023hierarchical,ha2024umilegs,liu2024visual}. 
Other approaches exploit the robot body or legs to extend its physical interaction capabilities~\cite{he2025visualleg,cheng2023legs}.
These capabilities provide an execution foundation for mobile manipulation. An additional task-level problem is deciding when to use physical interaction to resolve an obstruction and how to revise that decision when the observed response differs from the initial expectation. This work addresses this problem by coordinating navigation, obstacle pushing, and target manipulation, with persistent interaction evidence informing subsequent decisions. The contribution lies in task-level planning and obstacle memory rather than the development of a new whole-body controller.

\subsection{Navigation Among Movable Obstacles}

Navigation among movable obstacles (NAMO) studies how a robot can reach a goal by moving obstacles that block its path~\cite{stilman2005navigation,okada2004environment,stilman2008planning}. Classical approaches typically use geometric or physical models to reason about obstacle displacement and route feasibility with mapping\cite{wang2024prior,song2025map}. More recent methods incorporate visual perception, affordance estimation, and learning-based interaction strategies~\cite{bi2025interactive,wang2020affordance,ellis2022navigation,yao2024local}. Physical obstacle displacement has been explored using different robot interaction capabilities~\cite{bi2025interactive}. Recent work also introduces LLM-guided sampling-based search for NAMO~\cite{zhang2025namo}.
MoMa-MO builds on the idea of modifying the environment to obtain access, with the objective of completing a target manipulation task. Obstacle handling must therefore support access to feasible grasping and placement poses as the task progresses. Whereas the LLM-guided NAMO approach uses language-model reasoning to guide sampling-based search~\cite{zhang2025namo}, the proposed framework generates symbolic navigation and manipulation instructions conditioned on current task context and persistent obstacle records. Pushing outcomes update these records to guide subsequent obstacle handling.

\subsection{LLM-Based Task Planning}

LLM-based robotic planning uses language reasoning to support navigation and manipulation decisions. In vision-and-language navigation, LLMs support instruction interpretation and action selection using visual and contextual information~\cite{zhou2024navgpt,zhou2024navgpt2}, while map-based context enables spatial reasoning and multi-step planning beyond local observations~\cite{zhang2024mapgpt}. For manipulation in clutter, vision-language reasoning guides object selection and grasping, including the removal of obstructing objects to expose a target~\cite{qian2024thinkgrasp}. LLM planners also coordinate reusable visuomotor skills for long-horizon quadruped mobile manipulation~\cite{qiu2025wildlma}. To support execution in changing environments, dynamic spatio-semantic memory updates object representations across observations and enables localization as objects move, appear, or disappear~\cite{liu2025dynamem}.

\begin{figure*}[t]
    \centering
    \includegraphics[width=0.96\linewidth]{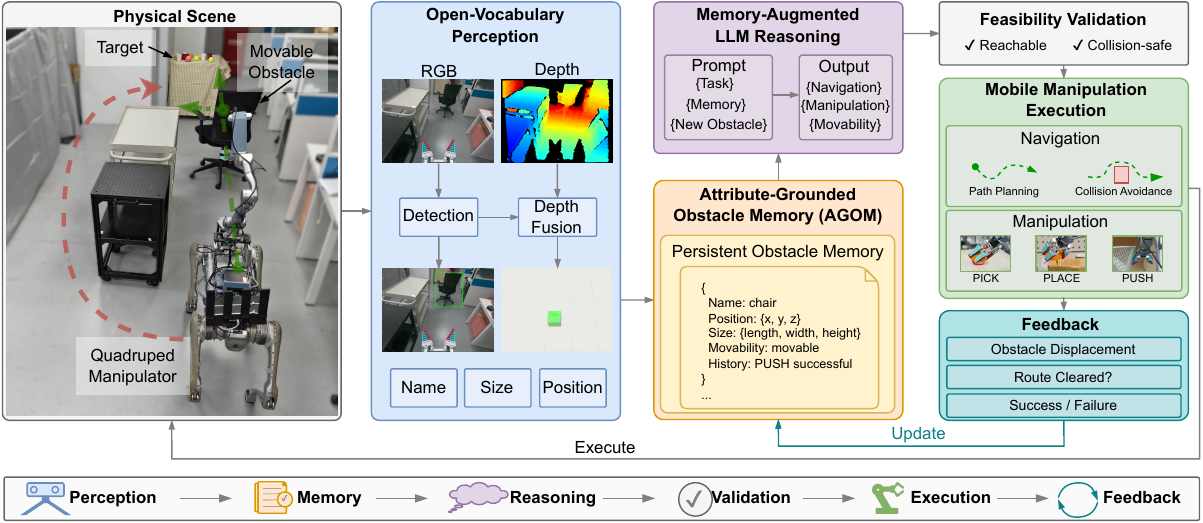}
    \caption{System overview. Perception converts sensor observations into textual scene information and AGOM records. An LLM generates move, pick, place, and push instructions, which are parsed, validated, and executed sequentially. Post-push observations update AGOM and inform subsequent planning.}
    \label{fig:sysover}
    \vspace{-2mm}
\end{figure*}

\section{Method}
\label{sec:method}

\subsection{Problem Formulation and System Overview}
\label{subsec:problem}

This work considers Mobile Manipulation among Movable Obstacles (MoMa-MO) for a quadruped manipulator performing a target pick-and-place task in an environment containing potentially movable obstacles. The robot must reach a feasible manipulation pose, grasp the target, transport it to a designated location, and complete placement. When an obstacle blocks the navigation route or target approach, the robot may either navigate around it or push it aside. In the current framework, obstacle handling is restricted to pushing, while picking and placing are reserved for the target task. The framework uses four symbolic command types:
\begin{equation}
\mathcal{U}
=
\{\mathtt{move},\mathtt{pick},\mathtt{place},\mathtt{push}\}.
\label{eq:primitive_types}
\end{equation}

Their interfaces are $\mathtt{move(pos)}$, $\mathtt{pick(name/pos)}$, $\mathtt{place(name/pos)}$, and $\mathtt{push(name/pos)}$, where $\mathtt{name/pos}$ denotes either a semantic category or a reference position. The LLM generates an ordered instruction sequence over these commands, while the execution modules resolve object references and generate the corresponding robot motions.
As shown in Fig.~\ref{fig:sysover}, the proposed framework consists of open-vocabulary perception, Attribute-Grounded Obstacle Memory (AGOM), LLM-based task reasoning, feasibility validation, mobile manipulation execution, and execution feedback. Perception extracts semantic and geometric object attributes from RGB-D observations. AGOM associates these observations with persistent obstacle records and previous interaction outcomes. The LLM combines the task goal, current scene, route context, and relevant AGOM records to generate an instruction sequence. Each instruction is validated before execution, and post-execution observations update the scene and AGOM for continued execution or replanning.
The framework separates task-level reasoning from motion generation. The LLM selects navigation and manipulation commands, while dedicated modules handle reachability checks, collision checking, path planning, and robot control.

\subsection{Open-Vocabulary Perception}
\label{subsec:perception}

YOLOE~\cite{yoloe} detects task-relevant objects in RGB images and provides semantic labels and detection regions. Valid depth samples within each detection region are filtered and back-projected to estimate the object's three-dimensional position and visible extent. The estimated positions are transformed into a shared map frame using camera calibration and the robot pose estimate. Each observation is represented as:
\begin{equation}
o_{j,t}
=
(c_{j,t},p_{j,t},s_{j,t}),
\label{eq:object_observation}
\end{equation}
where $c_{j,t}$ denotes the object category, $p_{j,t}$ its estimated position, and $s_{j,t}$ its estimated size. These attributes are used for object association, command argument resolution, and geometric validation. Since depth measurements may observe only part of an object, the estimated size is treated as an observation-dependent approximation.

The perception results are converted into textual scene descriptions for LLM reasoning. RGB-D images are not directly provided to the LLM. LiDAR and proprioceptive measurements are used separately for localization, occupancy mapping, and navigation.

\subsection{Attribute-Grounded Obstacle Memory}
\label{subsec:agom}

AGOM maintains persistent obstacle-specific information so that interaction outcomes can be reused during subsequent MoMa-MO decisions. It associates current semantic and geometric observations with obstacle identities, movability estimates, and interaction histories.

\paragraph{Persistent object records}
At time $t$, AGOM is represented as:
\begin{align}
\mathcal{M}_t &= \{m_{i,t}\}_{i=1}^{K_t},
\label{eq:memory}\\
m_{i,t} &= (id_i,c_i,p_{i,t},s_{i,t},\eta_{i,t},H_{i,t}),
\label{eq:memory_entry}
\end{align}
where $id_i$ is the persistent obstacle identity, $c_i$ is the semantic category, $p_{i,t}$ and $s_{i,t}$ denote the current position and size, $\eta_{i,t}$ is the movability estimate, and $H_{i,t}$ stores the interaction history. Each history entry records the attempted action, execution status, and observed outcome.
New observations are associated with existing records using semantic and geometric consistency. Matched observations update the current attributes while preserving identity and interaction history, whereas unmatched observations create new records. After a push, association accounts for possible obstacle displacement. If the association is ambiguous, the outcome is not assigned until the obstacle is reliably re-observed.

\paragraph{Movability estimation}
The movability estimate indicates whether an obstacle is expected to move under the available pushing action. The LLM initializes this estimate from the obstacle category, estimated position, and size. These observations provide useful cues but do not directly reveal physical properties such as loading or friction.
Interaction feedback is therefore used to revise the estimate during execution. For example, an obstacle predicted to be movable may remain stationary because of hidden load or environmental constraints. AGOM records the observed outcome together with the attempted push so that later decisions can account for this evidence.
A failed push is treated as evidence about the attempted interaction rather than proof that the obstacle cannot move under any configuration. Similarly, observable displacement does not imply that the obstacle has moved far enough to provide task access.

\paragraph{Task-relevant memory context}
Only obstacle records relevant to the current route or target approach are supplied to the LLM, together with the most recent interaction target. Each selected record contains the obstacle identity, category, geometry, movability estimate, and interaction history.
Route relevance is recomputed from the current task and scene rather than stored as a permanent obstacle property. An obstacle may therefore remain in AGOM after it no longer affects the current route, while its previous interaction evidence can be reused if it becomes relevant again later in the task.

\subsection{LLM-Based Task Reasoning}
\label{subsec:decision}

At planning step $t$, the LLM receives the task goal $g$, robot state $x_t$, current perception summary $\mathcal{S}_t$, route context $\Gamma_t$, and relevant AGOM records $\mathcal{M}^{\mathrm{rel}}_t$. It generates a symbolic instruction sequence:
\begin{align}
\hat{\Pi}_t
&=
\pi_{\mathrm{LLM}}
\left(
g,
x_t,
\mathcal{S}_t,
\Gamma_t,
\mathcal{M}^{\mathrm{rel}}_t
\right),
\label{eq:llm_policy}\\
\hat{\Pi}_t
&=
(\hat{u}_{t,1},\ldots,\hat{u}_{t,J_t}),
\label{eq:instruction_sequence}
\end{align}
where $J_t$ is the number of instructions and each $\hat{u}_{t,k}$ specifies a symbolic command and its argument. All inputs are serialized as text.

The prompt provides the current task state, perceived obstacles, available navigation and manipulation capabilities, route context, and relevant interaction history. The route context describes whether an obstacle blocks the current approach and whether an alternative navigation route is available. Based on this information, the LLM generates a task-level sequence that may include navigation, obstacle pushing, target picking, and target placement.
The movability estimate and interaction history support the decision between navigation and pushing. A feasible short detour may favor navigation, whereas a blocked or inefficient route may favor pushing an obstacle that is expected to move. If a previous push produces negligible displacement, the corresponding interaction history can instead favor an available detour. If the obstacle moves but route clearance remains insufficient, the updated scene triggers another planning step.
The LLM generates only symbolic commands and high-level arguments. Joint trajectories, contact forces, pushing directions, and controller references are determined by the execution modules. A pushing strategy may also contain a preceding move command for obstacle approach; therefore, the obstacle-handling decision is determined from the complete instruction sequence.

\subsection{Feasibility Validation}
\label{subsec:validation}

Before execution, each generated instruction is parsed and its arguments are resolved using the current perception results and AGOM records. Category-based references are matched to object instances, while position-based references specify object or destination locations. References to stored obstacles are resolved using their persistent identities and updated positions.

Each command is then checked for command validity, task-state compatibility, reachability, and collision constraints. A move command requires a feasible navigation route. Pick and place commands require valid gripper states and reachable manipulation configurations. A push command requires a reachable contact configuration and sufficient geometric clearance for the intended robot and obstacle motion. Intended obstacle contact during pushing is distinguished from unintended collision.
The navigation and manipulation modules generate candidate motions for these checks. Only validated commands are forwarded for execution. If validation fails, the corresponding failure information is returned for replanning.
Because validation relies on estimated geometry and motion models, it cannot guarantee successful physical execution. Localization error, incomplete observations, and uncertain obstacle response may cause the actual outcome to differ from the prediction. The remaining instruction sequence is therefore re-evaluated using the updated scene after execution.

\subsection{Mobile Manipulation Execution}
\label{subsec:execution}

\paragraph{Navigation}
LIO-SAM~\cite{lio-sam} provides robot pose estimates and registered LiDAR data for occupancy mapping. An $A^*$-based planner~\cite{astar} generates collision-free paths, which are tracked by a pure-pursuit controller~\cite{macenski2023regulated}. Navigation is used for explicit move commands and for approaching the poses required by pick, place, and push commands. Obstacles remain represented in the navigation map until updated observations indicate that the corresponding space has become free.

\paragraph{Target manipulation}
Pick and place commands consist of approach, grasp or release, and retreat motions. An MPC-based controller~\cite{mastalli2020crocoddyl} tracks the desired motion subject to the robot's kinematic and control constraints. If the target or placement region is outside the current manipulation workspace, the robot first navigates to a suitable pose.

\paragraph{Obstacle pushing}
For a push command, the execution module constructs a candidate pushing configuration:
\begin{equation}
\phi_{i,t}
=
(p^c_{i,t},\mathbf{d}_{i,t},\ell_{i,t}),
\qquad
\|\mathbf{d}_{i,t}\|_2=1,
\label{eq:push_configuration}
\end{equation}
where $p^c_{i,t}$ is the contact point, $\mathbf{d}_{i,t}$ is the pushing direction, and $\ell_{i,t}$ is the commanded stroke length. These parameters are selected from the estimated obstacle geometry and surrounding free space to move the obstacle away from the blocked route. The candidate configuration is checked for reachability and collision constraints before execution.

The push command consists of approach, contact establishment, bounded pushing motion, and retreat. The manipulation controller tracks the commanded end-effector motion within its operating limits. Execution terminates after completion or when the controller reports a failure or limit condition. The actual obstacle displacement is determined from post-push observations rather than inferred from the commanded motion.
Commands are executed sequentially, with each command returning its execution status before the next command is processed. After pushing, the affected region is re-observed before task execution resumes.

\subsection{Execution Feedback and Replanning}
\label{subsec:feedback}

Execution feedback closes the MoMa-MO loop by determining both the physical outcome of obstacle interaction and its effect on task access.

\paragraph{Obstacle displacement and route clearance}
After reliable object association, obstacle displacement is measured in the shared map frame:
\begin{align}
\Delta p_i
&=
\|p_i^{\mathrm{post}}-p_i^{\mathrm{pre}}\|_2,
\label{eq:push_displacement}\\
z_i^{\mathrm{move}}
&=
\mathbb{I}[\Delta p_i>\epsilon_p],
\label{eq:push_outcomes}
\end{align}
where $\epsilon_p$ is selected above the typical position-estimation noise. If object association or localization is unreliable, the movement outcome is marked as unknown rather than unsuccessful.
Route clearance is evaluated separately using the updated obstacle geometry and navigation map. The framework checks whether the intended passage is sufficiently clear and whether a feasible route to the required manipulation approach pose exists. An obstacle can therefore be classified as displaced while the corresponding route remains blocked. Similarly, a newly available detour is not interpreted as evidence that the push itself cleared the original route.

\paragraph{Memory update and replanning}
The observed interaction outcome is appended to the corresponding AGOM record together with the push command and controller status. Reliable post-push observations also update the obstacle geometry. The movability estimate is revised only when the interaction provides reliable evidence, including established contact, sufficient execution of the pushing action, and a valid displacement estimate.
When execution feedback invalidates the remaining plan, the updated scene and AGOM records are supplied to the LLM for replanning. The robot may proceed through the cleared route, select an available detour, attempt another validated interaction, or continue with the target manipulation task. Execution terminates if no feasible continuation is found or the task budget is exhausted.

Task completion is evaluated independently of obstacle displacement. A successful push therefore does not constitute successful MoMa-MO execution unless the robot subsequently completes the target pick-and-place task without unintended collision.

\section{Experiments}
\label{sec:experiments}

\subsection{Experimental Setup and Metrics}
\label{subsec:experimental_setup}

\paragraph{Platform and tasks}
Experiments are conducted on a Unitree B2 quadruped equipped with a Z1 robotic arm, an RGB-D camera, and LiDAR. The robot performs target pick-and-place tasks in environments where potentially movable obstacles block the navigation route or target approach. Obstacle handling is restricted to pushing, while picking and placing are reserved for the target task. GPT-5.6 Sol~\cite{openai_models} is used for obstacle movability assessment and task-level instruction planning, with all inputs provided as text. The same model is used across the compared LLM-based methods.

\paragraph{Compared methods}
Navigation-Only treats obstacles as fixed constraints and relies on collision-free navigation and replanning. Ours uses the complete MoMa-MO framework with AGOM-supported LLM reasoning to generate \texttt{move}, \texttt{pick}, \texttt{place}, and \texttt{push} instructions. No Interaction Feedback retains current obstacle attributes and initial movability estimates but excludes previous push outcomes and feedback-updated movability estimates from the LLM input. Where applicable, all methods share the same perception and execution modules, collision constraints, and task budgets. AGOM is reset between independent trials and retained throughout each trial.

\paragraph{Metrics}
Task success rate (\textbf{SR}) measures completion of the target pick-and-place task within the task budget and without unintended collision. Navigation success rate (\textbf{NSR}) measures arrival at a valid target approach pose within the task budget and without unintended collision, including access obtained through obstacle pushing. Intended pushing contact is not counted as collision. Completion time (\textbf{Time}) is measured from task initialization to successful placement, including processing and physical execution. Base travel distance (\textbf{Distance}) measures the total distance traveled by the robot base over the same interval. Time and Distance are reported as mean $\pm$ standard deviation over successful trials.

\subsection{Effect of AGOM on Movability Classification}
\label{subsec:agom_evaluation}

This experiment examines whether AGOM can use physical pushing feedback to correct initial LLM movability judgments during repeated observations. The dataset contains ten obstacle instances, including cases in which semantic expectations conflict with physical pushing outcomes, such as seemingly movable objects that resist pushing and seemingly immovable objects that can be displaced. Each instance is observed six times, yielding 60 observation records. At the first observation, the robot performs one pushing attempt to obtain feedback; subsequent observations involve no additional pushing. LLM Only independently classifies each observation using current semantic and geometric attributes, without retaining history or using pushing feedback. LLM + AGOM maintains a shared memory throughout the sequence, updates obstacle records using pushing outcomes, and retrieves the corresponding records upon subsequent observations. Classification is evaluated against movability labels established from the pushing outcomes under the tested conditions. We report accuracy (Acc), movable precision (MP), movable recall (MR), immovable precision (IP), and immovable recall (IR) on two sets: All contains all 60 observations, while Last contains the final observation of each obstacle, yielding ten records.

As shown in Table~\ref{tab:agom_eval}, LLM Only achieves 60.00\% accuracy on both sets, with unchanged class-wise precision and recall, indicating no aggregate improvement from repeated observations without interaction feedback. LLM + AGOM increases accuracy across all observations to 93.33\% and improves precision and recall for both classes. On the final observation of each obstacle, all metrics reach 100.00\%. These results demonstrate that retaining physical pushing feedback enables correction of initial movability judgments when previously encountered obstacles are observed again. The findings concern reuse of interaction evidence for the tested instances rather than generalization to unseen obstacles.

\begin{table}[t]
    \centering
    \caption{Movability classification with and without AGOM.
    Acc: accuracy; MP: movable precision; MR: movable recall;
    IP: immovable precision; IR: immovable recall.}
    \label{tab:agom_eval}
    \scriptsize
    \setlength{\tabcolsep}{2pt}
    \renewcommand{\arraystretch}{1.15}
    \begin{tabular*}{\columnwidth}{@{\extracolsep{\fill}}llccccc@{}}
        \toprule
        Method & Set & Acc & MP & MR & IP & IR \\
        \midrule
        LLM Only
        & All (60)
        & 60.00\%
        & 40.00\%
        & 66.67\%
        & 80.00\%
        & 57.14\% \\

        LLM + AGOM
        & All (60)
        & \textbf{93.33\%}
        & \textbf{85.00\%}
        & \textbf{94.44\%}
        & \textbf{97.50\%}
        & \textbf{92.86\%} \\
        \midrule
        LLM Only
        & Last (10)
        & 60.00\%
        & 40.00\%
        & 66.67\%
        & 80.00\%
        & 57.14\% \\

        LLM + AGOM
        & Last (10)
        & \textbf{100.00\%}
        & \textbf{100.00\%}
        & \textbf{100.00\%}
        & \textbf{100.00\%}
        & \textbf{100.00\%} \\
        \bottomrule
    \end{tabular*}
\end{table}

\begin{table}[t]
    \centering
    \caption{Task success across obstacle instances.}
    \label{tab:obstacle_success}
    \resizebox{\columnwidth}{!}{
        \begin{tabular}{llcc}
            \toprule
            Obstacle & Color & Dimensions (m) & Task SR $\uparrow$ \\
            \midrule
            Chair 1 & Black & $0.60 \times 0.60 \times 1.00$ & 9/10 \\
            Chair 2 & White & $0.60 \times 0.60 \times 0.90$ & 9/10 \\
            Box 1 & Yellow & $0.30 \times 0.30 \times 0.40$ & 8/10 \\
            Box 2 & Yellow & $0.25 \times 0.50 \times 0.50$ & 10/10 \\
            Waste Bin & Blue & $0.60 \times 0.60 \times 0.70$ & 8/10 \\
            \midrule
            Overall & & & 44/50 \\
            \bottomrule
        \end{tabular}
    }
\end{table}

\begin{table}[t]
    \centering
    \caption{Task performance with and without an available detour.}
    \label{tab:main_results}
    \resizebox{\columnwidth}{!}{
        \begin{tabular}{llcccc}
            \toprule
            Method & Detour & SR $\uparrow$ & NSR $\uparrow$ & Time (s) $\downarrow$ & Distance (m) $\downarrow$ \\
            \midrule
            Navigation-Only & Unavailable & 0/10 & 0/10 & N/A & N/A \\
            \midrule
            Navigation-Only & Available & \textbf{9/10} & \textbf{10/10} & 66.3 $\pm$ 0.5 & 13.9 $\pm$ 0.1 \\
            Ours & Available & \textbf{9/10} & 9/10 & \textbf{45.2 $\pm$ 0.4} & \textbf{7.1 $\pm$ 0.1} \\
            \bottomrule
        \end{tabular}
    }
\end{table}

\begin{table}[t]
    \centering
    \caption{Adaptation after an ineffective push.}
    \label{tab:failure_aware}
    \resizebox{\columnwidth}{!}{
        \begin{tabular}{lccc}
            \toprule
            Method & Detour Selection $\uparrow$ & Repeated Push Proposal $\downarrow$ & Task SR $\uparrow$ \\
            \midrule
            No Interaction Feedback & 0/10 & 10/10 & 0/10 \\
            Ours & \textbf{10/10} & \textbf{0/10} & \textbf{10/10} \\
            \bottomrule
        \end{tabular}
    }
\end{table}

\subsection{Task Success Across Obstacle Instances}
\label{subsec:obstacle_success}

End-to-end MoMa-MO execution is evaluated across five obstacle instances: two chairs, two cardboard boxes, and a cylindrical waste bin, with ten independently reset trials per instance. A successful trial requires pushing the obstacle sufficiently far to clear the task-relevant route and completing the target pick-and-place task under the criteria defined in Section~\ref{subsec:experimental_setup}.

As shown in Table~\ref{tab:obstacle_success}, the framework succeeds in 44 out of 50 trials, corresponding to an overall task success rate of 88\%. The success rate across individual obstacles ranges from 8/10 to 10/10. These results evaluate the complete MoMa-MO pipeline, including obstacle perception, task-level reasoning, navigation, pushing, route-clearance verification, and target manipulation.

\subsection{Navigation-Only versus Push-Enabled Execution}
\label{subsec:overall_performance}

Navigation-Only and Ours are compared under matched initial configurations in which an obstacle blocks the direct route while a collision-free detour remains available. Navigation-Only follows the detour, whereas Ours pushes the obstacle to clear the direct route before continuing the target task.

As shown in Table~\ref{tab:main_results}, both methods achieve a task success rate of 9/10. Navigation-Only achieves an NSR of 10/10, compared with 9/10 for Ours. Over successful trials, Ours reduces mean completion time from 66.3\,s to 45.2\,s and mean base travel distance from 13.9\,m to 7.1\,m, corresponding to reductions of 31.8\% and 48.9\%, respectively. These results demonstrate improved task efficiency in the tested configuration while maintaining the same observed task success rate, although navigation success is lower.

\begin{figure*}[t]
    \centering
    \includegraphics[width=0.98\linewidth]{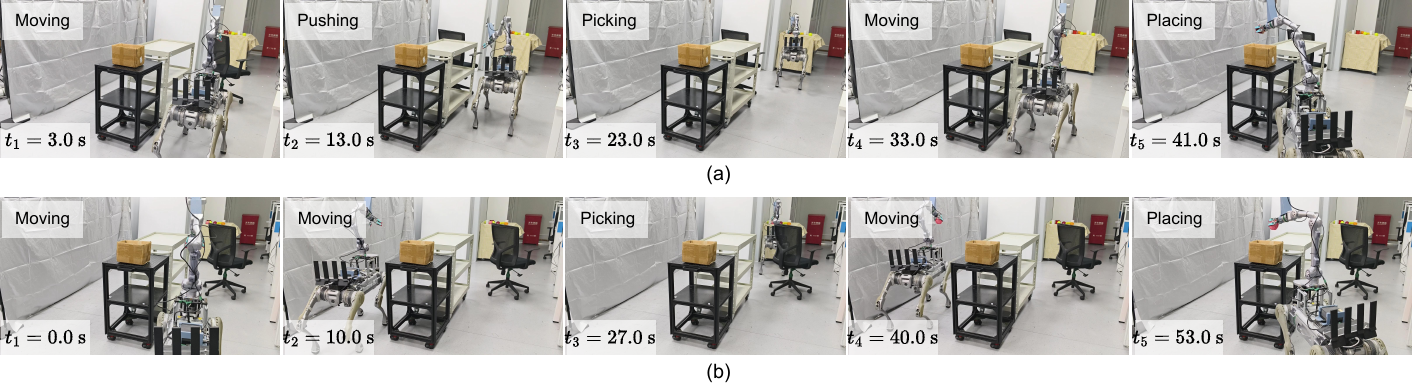}
    \vspace{-4mm}
    \caption{Representative executions with an available detour. (a) Ours pushes the obstacle to clear the direct route. (b) Navigation-Only follows a detour around the obstacle.}
    \label{fig:exp_comp_nav}
    \vspace{-4mm}
\end{figure*}

An additional evaluation tests Navigation-Only in a configuration without a feasible detour. Both NSR and SR are 0/10, indicating that navigation alone cannot provide task access in this configuration. This result characterizes the limitation of Navigation-Only and does not constitute a direct comparison with Ours, which is not evaluated under the same no-detour condition. Fig.~\ref{fig:exp_comp_nav} shows representative executions from the matched comparison with an available detour.

\begin{figure}[t]
    \centering
    \includegraphics[width=0.96\columnwidth]{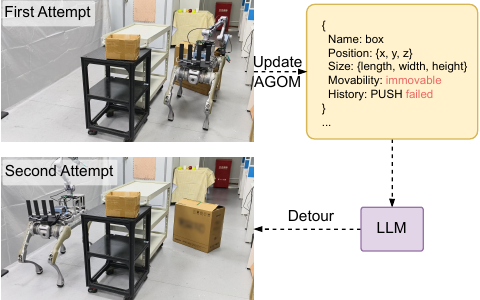}
    \caption{Adaptation after an ineffective push. AGOM records negligible displacement of a loaded box, and the subsequent plan selects the available detour.}
    \vspace{-2mm}
    \label{fig:failure_aware}
\end{figure}

\subsection{Adaptation after an Ineffective Push}
\label{subsec:failure_aware}

This experiment examines whether interaction feedback enables the framework to revise its obstacle-handling strategy after an ineffective push. A box containing a heavy load unknown to the robot blocks the direct route, while a feasible detour remains available. The initial push produces negligible obstacle displacement, providing negative evidence about the attempted interaction. Ours and No Interaction Feedback are evaluated from matched post-failure states with the same remaining task budget. Both methods receive the current perception and route information. No Interaction Feedback excludes the previous push outcome and feedback-updated movability estimate. Conversation history and feedback-derived summaries are also excluded from its LLM context to prevent indirect access to interaction feedback.

Performance is assessed using three metrics. Detour Selection measures whether the first post-failure instruction sequence selects the available detour without another push on the same obstacle. Repeated Push Proposal measures whether another push is proposed under unchanged conditions. Task SR measures successful completion of the remaining task. The proposal metrics are evaluated before feasibility validation or history-based rejection so that they reflect the LLM decision rather than execution-side intervention. As shown in Table~\ref{tab:failure_aware}, Ours selects the detour in all ten trials, proposes no repeated push, and completes all ten tasks. In contrast, No Interaction Feedback proposes another push in all ten trials and completes none within the remaining task budget. The results show that retaining interaction feedback allows the planner to revise its strategy when the observed obstacle response differs from the initial movability estimate. Fig.~\ref{fig:failure_aware} illustrates the transition from pushing to navigation.

\subsection{Computational Latency and Action Execution Time}
\label{subsec:runtime_analysis}

Table~\ref{tab:runtime_analysis} reports computational latency and action execution time separately. Detection and LLM latency are measured per invocation, controller computation is measured per control update, and action execution time is measured from command dispatch to completion.

The mean LLM reasoning latency is 3.05\,s. Since the LLM operates at the task-planning level rather than in the low-level control loop, this latency affects discrete planning events rather than high-frequency motion control. The mean execution times of \texttt{pick}, \texttt{place}, and \texttt{push} are 5.82\,s, 4.01\,s, and 8.96\,s, respectively. The entries in Table~\ref{tab:runtime_analysis} should not be summed to estimate total task duration because their invocation frequencies differ and processing may overlap with physical execution.

\subsection{Failure Modes}
\label{subsec:failure_modes}

\begin{figure}[t]
    \centering
    \includegraphics[width=0.95\columnwidth]{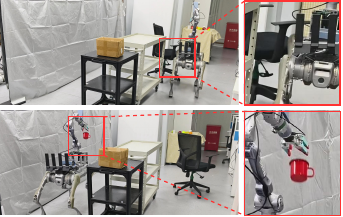}
    \caption{Representative failure modes. Top: insufficient route clearance causes unintended contact during passage. Bottom: the target object slips from the gripper before placement.}
    \label{fig:failure_mode}
\end{figure}

Two main failure modes are observed, as shown in Fig.~\ref{fig:failure_mode}. The first occurs when obstacle displacement is insufficient to provide route clearance, causing unintended contact during passage. Even if the robot subsequently completes the pick-and-place sequence, the trial is classified as unsuccessful under the collision-free task success criterion. This failure indicates that reliable route-clearance verification remains important after obstacle interaction.
The second failure mode is target-object slippage during navigation or manipulation, which causes task failure before placement. Slippage is observed more frequently during detour-based navigation. Longer travel distance and additional turning motions may contribute to this behavior, although their individual effects are not isolated in the current experiments.

\begin{table}[t]
    \centering
    \caption{Computational and action execution time.}
    \label{tab:runtime_analysis}
    \resizebox{\columnwidth}{!}{
        \begin{tabular}{llc}
            \toprule
            Component / Action & Measurement Basis & Mean Time (s) \\
            \midrule
            \multicolumn{3}{l}{\textit{Computational latency}} \\
            Open-Vocabulary Detection & Per invocation & 0.08 \\
            LLM Reasoning & Per invocation & 3.05 \\
            Controller Computation & Per control update & 0.01 \\
            \midrule
            \multicolumn{3}{l}{\textit{Action execution time}} \\
            \texttt{pick} & Per command & 5.82 \\
            \texttt{place} & Per command & 4.01 \\
            \texttt{push} & Per command & 8.96 \\
            \bottomrule
        \end{tabular}
    }
\end{table}

\section{Conclusion}
\label{sec:conclusion}

This work presents an LLM-guided framework for Mobile Manipulation among Movable Obstacles (MoMa-MO) on a quadruped manipulator. The framework combines task-level planning, feasibility validation, and execution feedback to coordinate navigation, obstacle pushing, and target manipulation. AGOM maintains persistent obstacle records containing semantic and geometric attributes, movability estimates, and interaction histories, allowing initial judgments to be revised using physical pushing feedback. Real-world experiments demonstrate end-to-end task execution across multiple obstacle instances, correction of movability judgments during repeated observations, and adaptation after ineffective pushes. In the matched comparison with an available detour, the framework maintains the same observed task success rate as Navigation-Only while reducing mean completion time by 31.8\% and mean base travel distance by 48.9\% over successful trials.

The current framework handles obstacles only through pushing, and observed failures highlight limitations in route-clearance verification and target-object retention during transport. Future work aims to incorporate additional obstacle interaction skills and improve clearance assessment and grasp stability in more complex environments.

\vspace{2mm}
\bibliographystyle{IEEEtran}
\normalem
\bibliography{References}

\end{document}